\documentclass[10pt,twocolumn,letterpaper]{article}

\usepackage{cvpr}
\usepackage{times}
\usepackage{epsfig}
\usepackage{graphicx}
\usepackage{amsmath}
\usepackage{amssymb}
\usepackage{algorithm}
\usepackage{algpseudocode}
\usepackage{authblk}

\usepackage[breaklinks=true,bookmarks=false]{hyperref}

\cvprfinalcopy % *** Uncomment this line for the final submission

\def\cvprPaperID{****} % *** Enter the CVPR Paper ID here

\begin{document}

%%%%%%%%% TITLE
\title{Dynamic Graph Generation Network: Generating Relational Knowledge from Diagrams}

%\author{Daesik Kim , Youngjoon Yoo, Jisoo Kim, Sangkuk Lee, Nojun Kwak \\
%Seoul National University\\
%V.DO Inc.\\

\author[1,2]{Daesik Kim}
\author[1]{Youngjoon Yoo}
\author[1]{Jeesoo Kim}
\author[1,2]{Sangkuk Lee}
\author[1]{Nojun Kwak}
\affil[1]{Seoul National University}
\affil[2]{V.DO Inc.}
\renewcommand\Authands{ and }


\maketitle
%\thispagestyle{empty}

%%%%%%%%% ABSTRACT
\begin{abstract}
   In this work, we introduce a new algorithm for analyzing a diagram, which contains visual and textual information in an abstract and integrated way. Whereas diagrams contain richer information compared with individual image-based or language-based data, proper solutions for automatically understanding them have not been proposed due to their innate characteristics of multi-modality and arbitrariness of layouts. To tackle this problem, we propose a unified diagram-parsing network for generating knowledge from diagrams based on an object detector and a recurrent neural network designed for a graphical structure. Specifically, we propose a dynamic graph-generation network that is based on dynamic memory and graph theory. 
   We explore the dynamics of information in a diagram with activation of gates in gated recurrent unit (GRU) cells.
   On publicly available diagram datasets, our model demonstrates a state-of-the-art result that outperforms other baselines.
   Moreover, further experiments on question answering shows potentials of the proposed method for various applications. 
   \vspace{-3mm}
\end{abstract}

%%%%%%%%% BODY TEXT

\section{Introduction}

Within a decade, performances on classical vision problems such as image classification \cite{he2016deep}, object detection \cite{girshick2015fast,liu2016ssd}, and segmentation \cite{long2015fully} have been largely improved by the use of deep learning frameworks. 
Based on the great successes of deep learning for such low-level vision problems, a next step could be deriving semantics from images such as relations between objects. 
For example, to understand a given soccer scene more deeply, it would be very important not only to detect the objects in the image, such as players and a ball but also to figure out the relationships between the objects.

\begin{figure}[t]
\begin{center}
   \includegraphics[width=\linewidth]{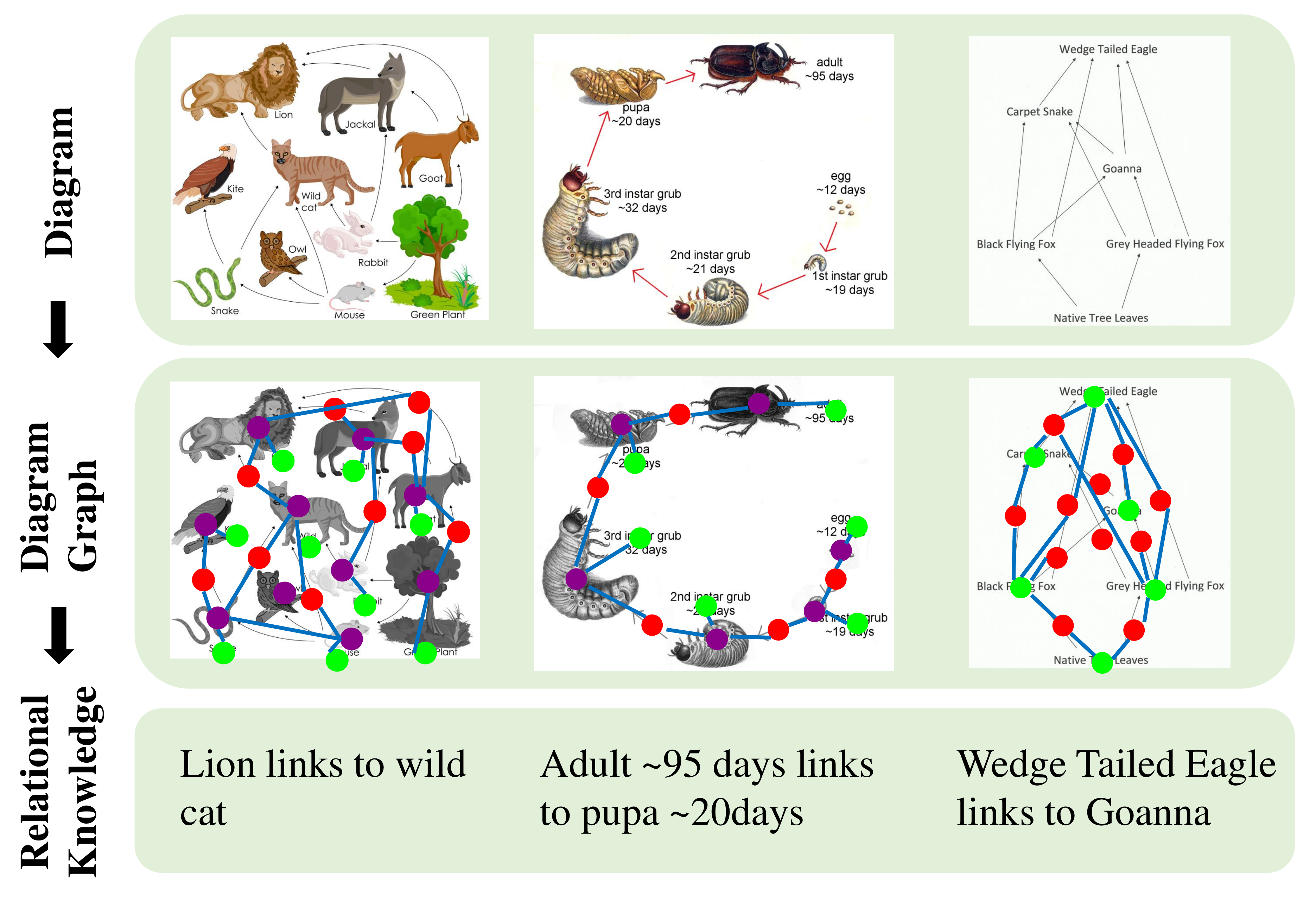}
\end{center}

   \caption{Examples of how relational knowledge can be generated from a diagram. In the first row, inputs are only diagrams which have various types of topics, illustrations, texts and layouts. Our model can infer a graphical structure in a diagram as in the second row. In the end, we can extract relational knowledge in the form of sentence from the generated graphs. 
   }
\label{fig:fig1}

\end{figure}

In this work, among various vision problems, we aim to understand diagram images, which have played a major role in classical knowledge representation and education.
Previously, most machine learning algorithms have focused on extracting knowledge from the information described by natural languages or structured databases (e.g. Freebase \cite{bollacker2008freebase},  Wordnet \cite{miller1995wordnet}).
In contrast to language-based knowledge, a diagram contains rich illustrations including text, visual information and their relationships, which can depict human's perception of objects more succinctly. 
As shown in Figure \ref{fig:fig1},  some complicated concepts such as ``food web in a jungle" and ``life cycle of a moth" can be easily described as a diagram. 
On the other hand, a single natural image or a single sentence may not be sufficient to deliver the same amount of information to the readers.

Whereas the diagram has good characteristics of knowledge abstraction, it requires composite solutions to properly analyze and extract the contained knowledge. 
Since diagrams in a science textbook employ a wide variety of methods for explaining concepts in their layout and composition, understanding a diagram can be a challenging problem of inferring human's general perception of structured knowledge. 
Unlike conventional vision problems, this task must involve inference models for vision, language and particularly relations among objects which can be a novel point. 
Despite the noted arbitrariness, we believe that a simple method generally exists to analyze and interpret the knowledge conveyed in a diagram.

There have not been many studies on diagram analysis yet, but Kembhavi~\etal~\cite{kembhavi2016diagram} recently proposed a pioneering work analyzing the diagram's structure.
The main flow of the algorithm is twofold:
1) Object detection: Objects in the diagram are detected and segmented individually by conventional methods such as those in \cite{arbelaez2014multiscale,kokkinos2010highly}. 
2) Relation inference: The relations among detected objects are inferred by a recurrent neural network (RNN) to transmit contexts sequentially.
However, this approach has several limitations. 
First, concatenating separated methods results in a long pipeline from input to output, which can cause accumulated errors and lose contexts on a diagram. 
%The method detects and segments the objects individually and hence, \nj{can} not fully utilize the relation of \nj{contexts} on a diagram. 
Second, and more importantly, %while characteristics of relationships can be defined as a graph, 
the vanilla RNN is not fully capable of dealing with the information formed as a graph structure. 
%Thus, a vanilla RNN could miss useful properties of graph structures.

\begin{figure*}[t]
\begin{center}
\includegraphics[width=\linewidth]{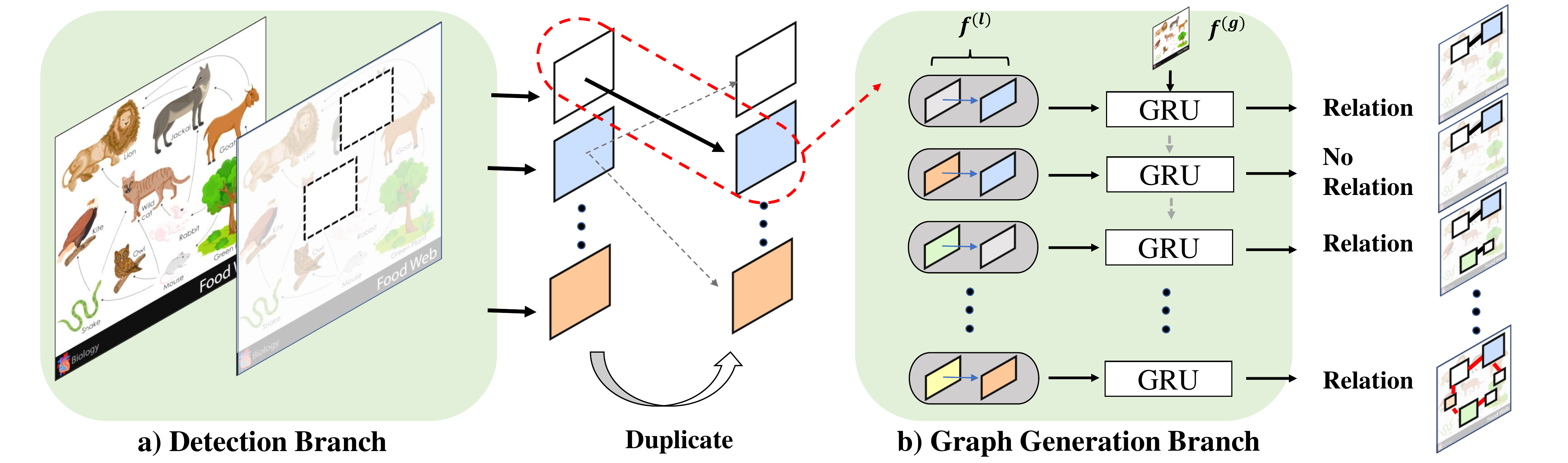}
\end{center}
\vspace{-3mm}
   \caption{Overview of the \textit{unified diagram parsing network} (UDPnet). In (a) the detection branch, an object detector can extract $n$ objects of 4 different types. Then in order to exploit pairs of objects, we produce $n^2$ relationship candidates with duplicated objects. (b) In the graph generation branch, we pass local features $f^{(l)}$ from $n^2$ candidates to the \textit{dynamic graph generation network} (DGGN) with a global feature $f^{(g)}$. In the final step, each relationship candidate can be determined whether it is valid or not. At last, we can establish a relationship graph with nodes and edges.}
\label{fig:long}
\vspace{-3mm}
\end{figure*}

In this paper, we propose a novel method to solve the aforementioned issues.
Our contributions are twofold. First, using a robust object detection model and a novel graph-based method, a \textit{unified diagram parsing network} (UDPnet) is proposed to understand a diagram by jointly solving the two tasks of object detection and relation matching, which tackles the first limitation of the existing work.
%The entire flow of generating relational knowledge from a diagram is described in Figure \ref{fig:long}.
Second, we propose a RNN-based \textit{dynamic graph generation network} (DGGN) to fully exploit the diagram information by describing with a graph structure. 
To solve the problem, we propose a dynamic adjacency tensor memory (DATM) for the DGGN to store information about the relationships among the elements in a diagram.
The DATM has features of both an adjacency matrix in graph theory and a dynamic memory in recent deep learning.
%To solve the problem, we propose a novel module in the DGGN, the dynamic adjacency tensor memory (DATM), which has features of both an adjacency matrix in graph theory and a dynamic memory in recent deep learning. 
With this new type of memory, the DGGN suggests a novel way to propagate information through the structure of a graph.
In order to demonstrate the effectiveness of the proposed DGGN, we evaluated our model on a couple of diagram datasets. We also analyzed the inside of GRU \cite{cho2014properties} cells to observe the dynamics of information in the DGGN.

% we make the following contributions:
%\begin{itemize}
%\item We propose a unified diagram parsing network(UDP net) for understanding a diagram by jointly solving two tasks: object detection and relation matching.
%\item We introduce a novel RNN with dynamic adjacency matrix memory to propagate information based on graph structure.
%\item We evaluate our model on two scientific diagram datasets for generating graph and question answering.
%\end{itemize}

%This paper is organized as follows. Section \ref{sec:relative_work} reviews the related works on object detection, visual relationships detection, neural network models on a graph and memory networks. Section \ref{sec:proposed_method} proposes our model of UDPnet, which consists of an object detection branch and a graph generation branch. In section \ref{sec:experiment}, our experimental setting and training details are described. Then the quantitative of the proposed method are presented. Moreover, analysis of DGGN and qualitative results are discussed in Section \ref{sec:discussion}.

\section{Related Works}
\label{sec:relative_work}

%\noindent \textbf{Object detection:} 
%In early days, the sliding window method was applied to object detection problems with hand-crafted features. 
%With the success of deep learning, two-stage approaches such as the family of R-CNN \cite{girshick2014rich,girshick2015fast,ren2015faster} became a dominant paradigm in modern object detection. In this family, the first stage proposes candidate RoIs (region of interests) from the backbone CNN such as VGGnet, then the second stage classifies foreground and background classes with a fully-connected network. 
%Fast R-CNN \cite{girshick2015fast} introduced the concept of an RoI pooling which utilizes shared feature maps for different region proposals. Faster R-CNN \cite{ren2015faster}, a later model, adopted region proposal network (RPN) which integrated box proposal and classifier. 

%Recently, several single stage approaches have been suggested due to their simplicity and efficiency. For instance, SSD \cite{liu2016ssd} and YOLO \cite{redmon2016you} were implemented with simple architectures and showed better speed compared to the two-stage models \nj{by sacrificing a little accuracy}. Particularly, SSD only consists of convolution and pooling layers, localizing and classifying objects from feature pyramids. 
%\nj{Other} extensions of SSD \cite{ren2017accurate,fu2017dssd, Jungjisoo} have been proposed to overcome \nj{the} weakness of accuracy. Our method is based on SSD frameworks.

\noindent\textbf{Visual relationships:} 
Studies on visual relationships have been emerging in the field of computer vision. 
This line of research includes detection of visual relationships \cite{lu2016visual,li2017vip} and generation of a scene graph \cite{johnson2015image}. 
Most of these approaches are based on algorithms for grouping elements by relationships, and aiming to find relationships among the elements. 
%The problem of human-object interactions concerns actions of a human and interactions with objects, which requires a deeper understanding of human behaviors. 
%More generally, scene graph generation necessitates understanding of general scenes in natural images. 
Recently, this research field has focused on the scene graph analysis algorithm, which tackles the problem of understanding general scenes in natural images.  
Lu \etal \cite{lu2016visual} incorporated language prior to reasoning over a pair of objects and Xu \etal \cite{xu2017scene} solved scene graph inference using GRUs via iterative message passing.
Whereas most of the previous studies dealt with natural images, we aim to infer visual relationships and generate a graph based on these relationships.
Moreover, our method extracts knowledge from an abstracted diagram by inferring human's general perception. \\

\noindent\textbf{Neural networks on a graph:} Generalization of neural networks for arbitrarily structured graphs has drawn attention in the last few years. %Most approaches are based on RNN or CNN which are generalized well in practice. 
\textit{Graph neural networks} (GNNs) \cite{scarselli2009graph} were introduced as an RNN-based model that iteratively propagates nodes in the graph until the nodes reach a stable fixed point. 
Later, Li \etal \cite{li2015gated} proposed \textit{gated graph neural networks} (GG-NNs), which apply GRU as an RNN model for the task. 
%which adapted modern practices for RNN such as GRUs to exploit gates and backpropagation through time (BPTT) in computing gradients. 
In contrast to the RNN-based models, Marino \etal \cite{marino2016more} proposed \textit{graph search neural network} (GSNN) to build knowledge graphs for multi-label classification problems. 
GSNN iteratively predicts nodes based on current knowledge by a way of pre-trained `importance network'. 
The main difference between previous methods and ours is that our model can generate a graph structure based on relationships between nodes. 
Since generating a graph should involve in dynamic establishment or removal of edges between nodes, we also adopt RNN for DGGN as most neural-network-based methods for a graph. 
%
%can be whether the structure of a graph is already established or built online during both training and inference. 
The proposed DGGN not only works by message-passing between nodes, but also builds the edges of a graph online, which provides great potential for graph generation and solution of inference problems.    
\\

\noindent \textbf{Memory augmented neural network:} Since Weston \etal \cite{weston2014memory} proposed a memory network for the question answering problem, a memory augmented network became popular in natural language processing. 
This memory component has shown great potential to tackle many problems of neural networks such as catastrophic forgetting. 
In particular, Graves \etal applied the memory component in Neural Turing machine \cite{graves2014neural}, and showed that the neural network can update and retrieve memory dynamically.
%by using a supportive storage of representations. 
Using this concept, a number of dynamic memory models have been proposed to solve multi-modal problems such as visual question answering \cite{kumar2016ask,xiong2016dynamic}. 
In this paper, we incorporate the scheme of the dynamic memory in DGGN to easily capture and store the graph structure.
\\

\section{Proposed Method}
\label{sec:proposed_method}

%In our method, the diagram understanding problem is tackled in two steps; 1) detecting a set of objects $O = \{o_i\}_{i=1}^n$, and 2) finding a set of relations $R = \{r_j\}_{j=1}^m$ from a diagram image. Specifically, each of $n$ objects, $o_i$, is described \js{as} $<location, class>$, and the relation $r_j$ \js{is in the form of} $<o_1,o_2>$. 
%To achieve the goal, we should detect objects on a diagram and decide whether a pair of objects has a relation or not, which is not a straightforward problem. 

Figure \ref{fig:long} shows a overall framework of the proposed UDPnet.
The proposed network consists of two branches: 
1) an object detection network, 
and 2) a graph generation network handling the relations among the detected objects.
In the first branch, a set of objects $O = \{o_i\}_{i=1}^n$ in a diagram image is detected.
In the second branch, the relations $R = \{r_j\}_{j=1}^m$ among the objects are generated.
We define an object $o_i$ as $<location, class>$, and a relation $r_j$ in the form of $<o_1,o_2>$. 
Both branches can be optimized simultaneously by a multi-task learning method in an end-to-end manner.
After the optimization process, we can use the generated relational information to solve language-based problems such as question-answering.

\subsection{Detecting Constituents in a Diagram}

As seen in the Figure \ref{fig:fig1}, various kinds of objects can be included in a diagram depending on the information being conveyed. 
%For example, in diagrams dealing with scientific subjects, many species of animals and planets are drawn in the form of illustration. 
Those objects are usually described in a simplified manner and the number of object classes is huge, which makes detecting and classifying objects more difficult. 
In our work, instead of detecting classical object types such as cats and dogs, we define objects in four categorical classes which are adequate for diagrams: blob (individual object), text, arrow head and arrow tail. 
As a detector, we used SSD \cite{liu2016ssd} which has been reported to have a robust performance.

\subsection{Generating a Graph of relationships }

\subsubsection{Overall Procedure of Graph Generation}

In our method, the relation matching for objects in a diagram is conducted by predicting the presence of an edge between a pair of vertices using graph inference.
The nodes and edges of a graph match to the objects and the relations of paired objects, respectively.
Therefore, the graph is described as a bipartite graph, 
\begin{equation} 
G=(V,E),
\label{eq:bipartite_graph}
\end{equation}
where $V = X \cup Y$ represents the set of paired disjoint vertices $X\subset V$ and $Y\subset V$, and $E$ denotes edges of the graph each of which connects a pair of nodes $x \in X$ and $y \in Y$.
To construct a bipartite graph, we duplicate the detected objects $O$ as $O_x$ and $O_y$ and assume that those two sets are disjoint.
Then we predict whether an edge between the nodes $o_x \in O_x$ and $o_y \in O_y$ exists.

The connection between nodes is determined by their spatial relationship and the confidence score for each object class which is provided by the object detector. 
Note that we do not use convolution features from ROI pooling because there can be various kinds of objects in a diagram, whose shape and texture are hard to be generalized.
Instead, we define a feature $f_x \in \mathbb{R}^{13}$ for the object $o_x$ including location (xmin, ymin, xmax, ymax), center point (xcenter, ycenter), width, height and confidence scores. 
Thus, the relationship between two objects $o_x$ and $o_y$ is described as local feature $f^{(l)} = [f_x, f_y] \in \mathbb{R}^{26}$, and the
feature vector $f^{(l)}$ acts as an input to a RNN layer.
% To achieve greater flexibility in a more principled training framework, we use a generic RNN unit instead, in particular a Gated Recurrent Unit (GRU) [7].
 %and a sequence of \sout{$f^{(l)}$} \js{these vectors} is used to predict the existence of edges iteratively. 
To prevent the order of local features in a sequence from affecting the performance,
%To remove the effect of \nj{different order of local features input GRU}, 
we randomly shuffle the order of the features before training in every iteration.

Furthermore, to extract the layout of a diagram and spatial information of all objects, a global feature $f^{(g)}$ is utilized as an input to the RNN.
%, while local features can hardly contain information on the contexts. 
The global feature $f^{(g)}\in \mathbb{R}^{128}$ is constructed by the sum of the convolution feature of conv-7 layer ($256 \times 1 \times 1 $) of backbone network in the first branch and the binary mask feature of a diagram ($128 \times 1 $). 
To match the dimension of conv-7 feature as that of hidden units, we use a fully connected layer in the last step.
For the mask feature, we pass the $\mathbb{R}^{n_h \times n_w \times n_c}$ dimensional binary mask map to the 4 layered convolution and max pooling to match the dimension to the hidden unit, where $n_h$ and $n_w$ are the width and height of an image, and $n_c$ is the number of object classes.

%The mask feature is obtained as the following: 1) create a zero tensor $T_m\in \mathbb{R}^{n_h \times n_w \times n_c}$ where $n_h$ and $n_w$ are the width and height of an image which are identically set to be 300 in our experiments as we used SSD-300 model, $n_c$ is the number of object classes which is 4 in our model, 2) add 1 in pixels on the location of positive objects and 3) pass \js{it} through 4 layers of convolution and max-pooling to match the dimension with that of hidden units ($128 \times 1$). These attached layers are able to be trained with the backbone network.

\subsubsection{Dynamic Graph Generation Network}

\begin{figure}[t]
\begin{center}
   \includegraphics[width=\linewidth]{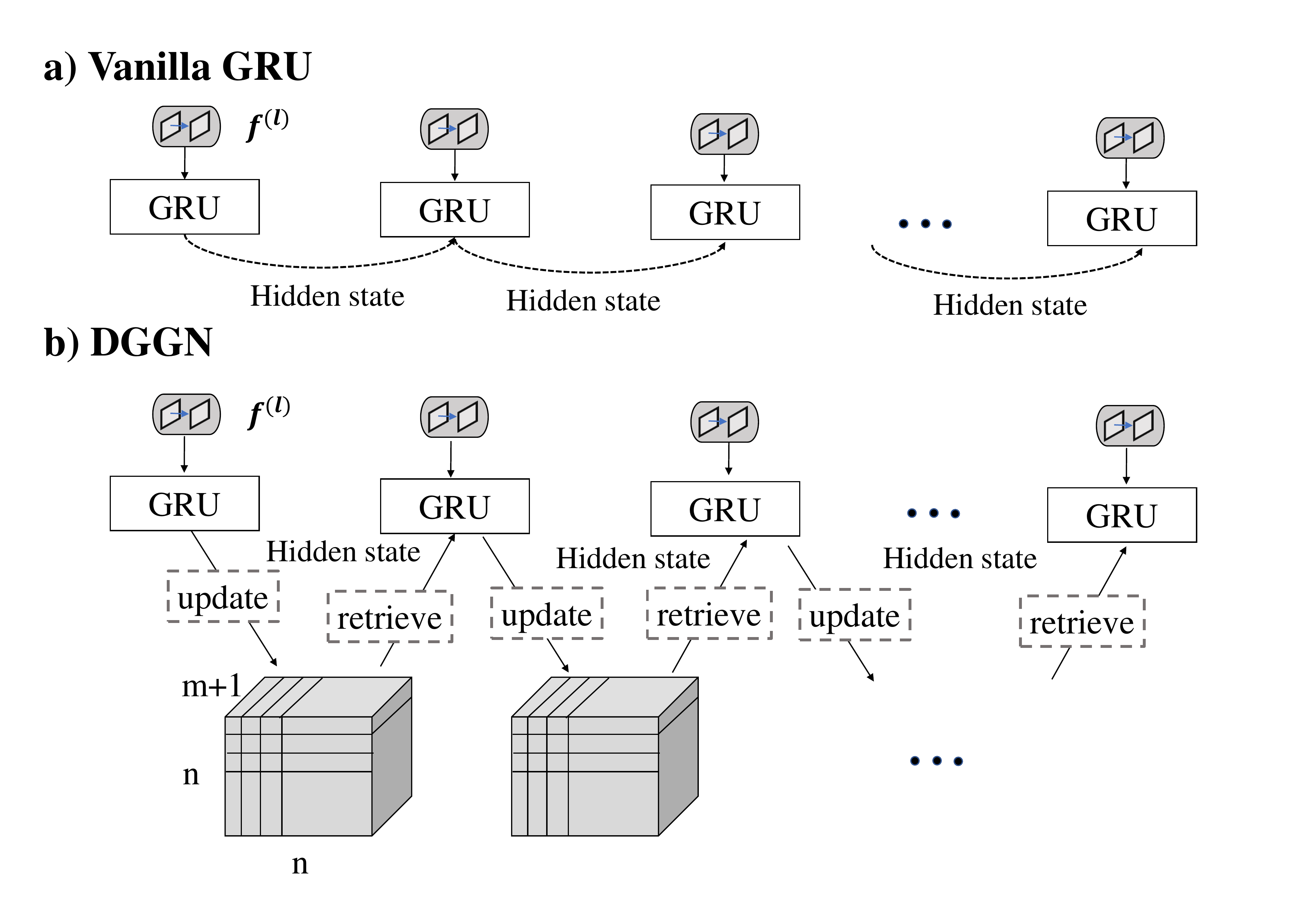}
\end{center}
\vspace{-2mm}
   \caption{Comparison of the vanilla GRU and the proposed DGGN. (a) In vanilla RNN, information is sequentially transmitted only to a randomly selected next cell. (b) In DGGN, past hidden states are calculated with the dynamic adjacency memory, and the information on the entire graph is propagated in both the update and the retrieval processes simultaneously.
   }
\label{fig:GRU_comparison}
\vspace{-2mm}
\end{figure}

In our problem, the local feature vector $f^{(l)}_{i,j}, (i,j = 1,...,n)$ contains the connection information between the nodes $o_i\in X$ and $o_j\in Y$. 
For simplicity, instead of two indices $i$ and $j$, we will use one index $t$ to denote the local feature, \ie  $f^{(l)}_{t}, (t= 1,...,n^2)$. 
In the previous work \cite{kembhavi2016diagram}, vanilla RNN was used and the connection vector $f^{(l)}_t$ was inputted sequentially to train the RNN. 
The problem is that there is no guarantee that the input $f^{(l)}_t$ will be associated with the $f^{(l)}_{t+1}$ because the vector $f^{(l)}_t$ is randomly shuffled in stochastic gradient training. 
Besides, while we define this problem as the bipartite graph inference, vanilla RNN could not capture the graph structure and propagate it into the next unit. 

To solve the aforementioned problem, we propose the DGGN method which incorporates GRU as a base model.
As shown in Figure \ref{fig:GRU_comparison}, the proposed method of propagating previous states to the next step is completely different from that of the vanilla GRU. 
In order to exploit the graph structure, instead of just sequentially transferring features as in vanilla GRU (Figure \ref{fig:GRU_comparison}(a)), we aggregate messages from adjacent edges (Figure \ref{fig:GRU_comparison}(b)).
%In contrast to the vanilla GRU which already has a defined structure, 
To pass the messages from adjacent edges, the proposed DGGN requires a dynamic programming scheme which can build the graph structure in an online manner.

\begin{figure}[t]
\begin{center}
   \includegraphics[width=\linewidth]{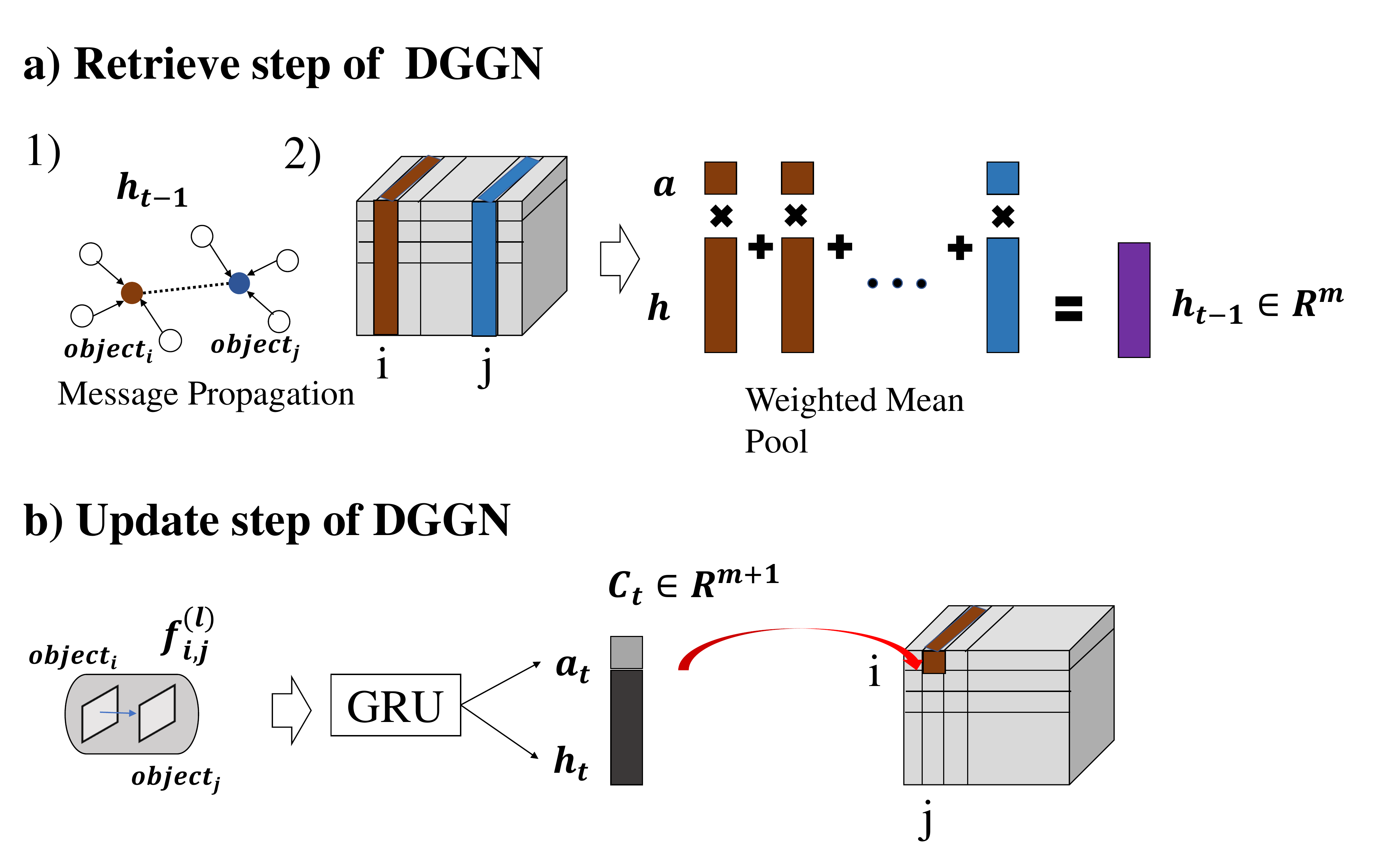}
\end{center}
\vspace{-2mm}
   \caption{Specific explanations of \textit{update} and \textit{retrieve} steps with the DATM in DGGN. (a) In the retrieve step, past messages are transmitted from adjacent edges (a-1). Specifically, to obtain previous hidden state, we conduct weighted mean pool with extracted matrix at indexes of objects. (b) In the update step, model can store the inferred information into the DATM with a concatenated vector at indexes of input objects.}
\label{fig:dynamic_memory}

\end{figure}

In this paper, we incorporate the adjacency matrix in the graph theory which has been mainly used to propagate message through known structure of the graph~ \cite{scarselli2009graph}. 
However, in our problem, the adjacency matrix is unknown, which has to be estimated. 
Therefore, we propose a dynamic memory component into this problem which holds the connection information among nodes.
%among $f^{(l)}_t$. 
%
In this work, we expand $2$-dimensional adjacency matrix to $3$-dimensional memory. 
The \textit{dynamic adjacency tensor memory}(DATM) $D\in \mathbb{R}^{n \times n \times (m+1)}$ is defined as a concatenation of the adjacency matrix $A \in \mathbb{R}^{n \times n}$ and the corresponding hidden unit $H$ whose ($i,j$) element $h_{i,j}$ is an $m$ dimensional hidden vector of the GRU which is related to the connection between the nodes $o_i$ and $o_j$.
The adjacency matrix $A$ represents the connection status between each of $n$ nodes in the directed graph.
Each cell in the adjacency matrix only indicates whether the corresponding pair of nodes has a directed arc or not. 
Then both \textit{retrieve} and \textit{update} steps with tensor $D$ are implemented to aggregate messages from adjacent edges and to build up graph simultaneously. \\

\noindent \textbf{Retrieve Step:} Figure \ref{fig:dynamic_memory}(a) shows the \textit{retrieve step} of DGGN. 
We can get the previous hidden state $\hat{h}_{t-1}$ which collects messages propagated through adjacent edges (Figure~\ref{fig:dynamic_memory} (a-1)). 
In doing so, as shown in Figure~\ref{fig:dynamic_memory}(a-2) and equation (\ref{eq:damm_ht}), we take average of
the adjacent vectors of $o_i$ and $o_j$ weighted by the probability of the existence of an edge. Formally, we extract an adequate hidden unit $\hat{h}_{t}$ for the input vector $f^{(l)}_{t+1}$ representing the connection with node $i$ and $j$, as in %equation~(\ref{eq:damm_ht}),
\begin{equation} 
\label{eq:damm_ht}
\hat{h}_{t}=\sum^{n}_{k=1}a_{k,i}h_{k,i} + \sum^{n}_{k=1}a_{k,j}h_{k,j} + f^{(g)}.
\end{equation}
Here, $a_{i,j}$ represents the $(i,j)$ element of the matrix $A$, and $h_{i,j} \in \mathbb{R}^m$ is the hidden unit stored in the $(i,j)$ location of the tensor $H$. 
In this step, the probability $a_{i,j}$ works as weights for aggregating messages which represents the philosophy that more credible adjacent edges should give more credible messages.
Before transmitted to GRU layer, the global feature $f^{(g)}$ is added to reflect the global shape of the diagram. \\

\noindent \textbf{Update Step:} In the \textit{update step} shown in Figure~\ref{fig:dynamic_memory}(b) 
%equation (\ref{eq:gru1} -- \ref{eq:output})
, we update the cell $D_{ij}$ with an $m+1$ length vector that concatenates the output $a_t$ and the hidden state ${h}_t$ from a GRU cell (\ref{eq:output}). 
\begin{equation} 
\label{eq:gru1}
r_t = \sigma(W_{xr}f_t + W_{hr}\hat{h}_{t-1} + b_r),
\end{equation}
\begin{equation} 
\label{eq:gru2}
z_t = \sigma(W_{xz}f_t + W_{hz}\hat{h}_{t-1} + b_z),
\end{equation}
\begin{equation} 
\label{eq:gru3}
\bar{h}_t = \text{tanh}(W_{xh}f_t + W_{hh}(r_t\odot{}\hat{h}_{t-1}) + b_h),
\end{equation}
\begin{equation} 
\label{eq:gru4}
h_t = z_t\odot{}\hat{h}_{t-1} + (1 - z_t)\odot{}\bar{h}_t,
\end{equation}
\begin{equation} 
\label{eq:gru5}
a_t=\sigma(W_l h_t+b_l),
\end{equation}
\begin{equation} 
\label{eq:output}
D_{ij}=[a_t,h_t].
\end{equation}
Here, $\sigma(\cdot)$ is a sigmoid function. To obtain the hidden state $\hat{h}_t$, the vectors $\hat{h}_{t-1}$ and $f^{(l)}_{t}$ are used as previous hidden state and input vectors of the standard GRU, respectively. Update gate $z_t$ has a role to adjust influx of previous information $\hat{h}_{t-1}$ in the GRU cell (\ref{eq:gru4}). The binary output $a_t$ is obtained after fully connected layer (\ref{eq:gru5}).

\subsection{Multi-task Training and Cascaded Inference}

In this work, the proposed UDPnet shown in Figure~\ref{fig:long} is trained in an end-to-end manner.
Because the UDPnet consists of two branches (object detection by SSD and graph generation by DGGN), by nature, the problem is a multi-task learning problem. Thus, different losses for each branches are combined into the overall loss $L$ as follows: 
\begin{equation} 
L = \alpha L_c + \beta L_l + \gamma L_r.
\label{eq:loss}
\end{equation}
The overall loss is a weighted sum of the classification loss $L_c$ and the location regression loss $L_l$ for the object detection branch, and the relation classification loss $L_r$ for the graph generation network.

As defined in original SSD, the classification loss $L_c$ is the softmax loss over confidences of multiple classes and the location regression loss $L_l$ is a smooth L1 loss \cite{girshick2015fast} between the predicted box and the ground truth box. 
The relation classification loss $L_r$ is the softmax loss over two classes, adjacent or not. 
For a faster convergence, we first pre-trained object detection branch alone, then fine-tuned both branches jointly with the overall loss.

During training, matching strategy between the candidates and the ground truths is important for both box detection and relationship inference. 
To solve the issue, we set our own strategy for matching candidate pairs and the ground truth. 
First, given $n$ objects detected at the first branch of object detection, we generate $n^2$ pairs of relation candidates. 
For each relation candidate, the two intersection over unions (IOUs), each of which is computed between one of the detected objects and the closest ground truth object, are averaged. 
Then, each ground truth relationship is matched with the best overlapped relation candidate. 
%We set number of candidates and ratio of positive and negative samples. 
To consider the imbalance in the number of detected objects among different diagrams, 
%match the number of detected objects, 
we should sample the same number of relation candidates from each training diagram. 

At inference, we first detect objects in a diagram. 
Then we apply non maximum suppression (NMS) with an IoU threshold of 0.45 on boxes with scores higher than 0.01. 
Unlike in training, we should use all boxes that were detected to generate candidate pairs for next branch. 
Next, we apply graph generation branch to all relation candidates to infer relationship to each other. 
Finally, we can obtain a diagram graph composed of adjacent edges between nodes with confidence scores higher than 0.1. 

After graph inference, we can post-process the generated relational information to further generate knowledge sentences which can be inputs of question answering models. 
Thus, our methods can make a bridge between visual inference and linguistic reasoning.
Actually, we applied proposed pipeline in this paper to Textbook Question Answering competition \cite{Kembhavi2017tqa} and the details on the post-processing can be found in the supplementary material.

%---------------------------------------
\begin{figure*}[t]
\begin{center}
   \includegraphics[width=\textwidth]{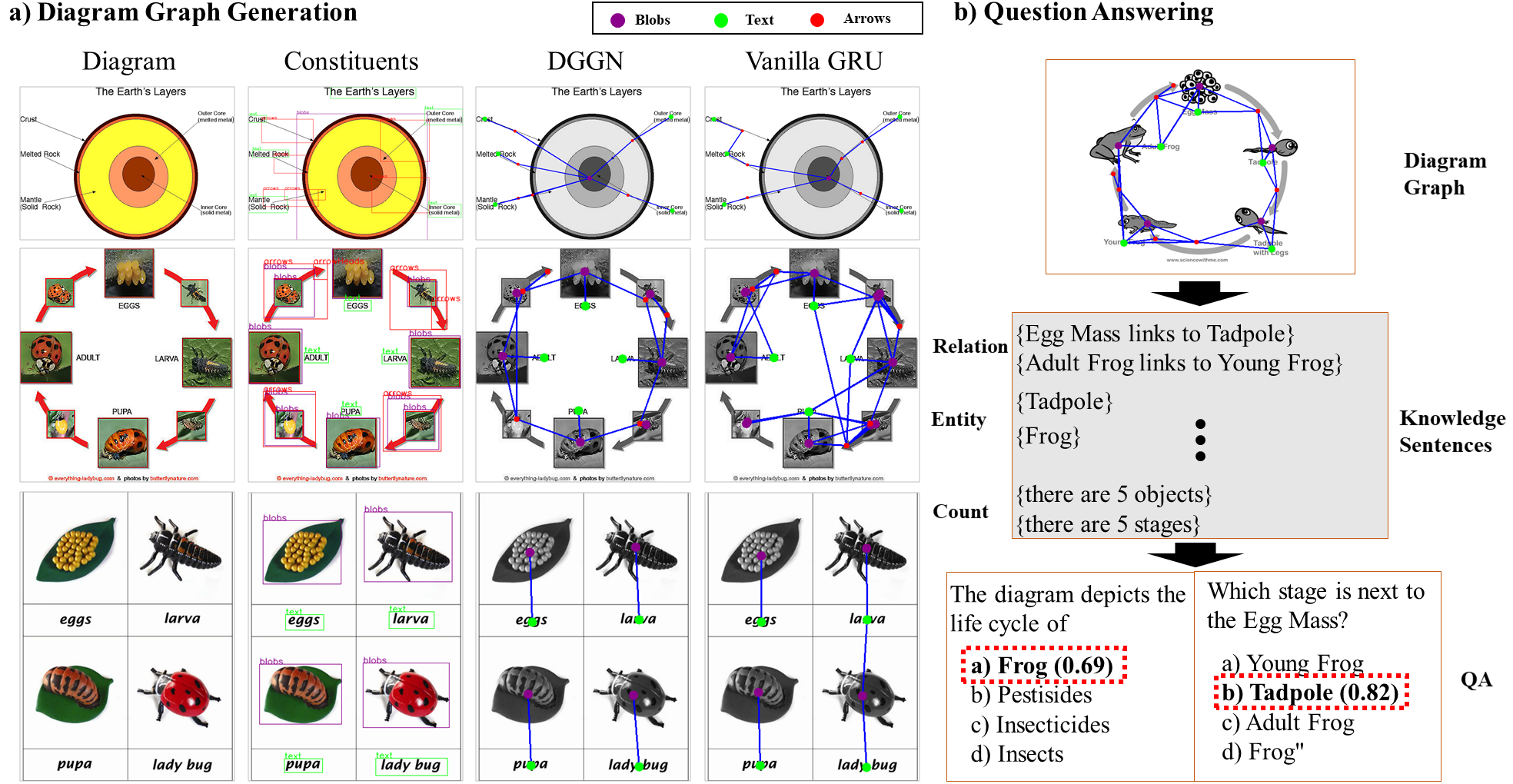}
\end{center}
\vspace{-2mm}
   \caption{Qualitative results of diagram graph generation and a pipeline to solve question answering problem. (a) Each row shows an example of various kinds of diagram. From the left, original diagrams and diagrams with detected constituents are presented. In last two columns, comparison between the DGGN and the Vanilla GRU with final results is shown. (b) From a diagram graph, we extract knowledge sentences, then solve multi-choice problems. }
\label{fig:fig7}
\vspace{-2mm}
\end{figure*}
%---------------------------------------

\section{Evaluation}
\label{sec:experiment}

In this section, we validate the performance of the proposed 
algorithm for the two sub-problems: graph generating and question answering.
\\

%\subsection{Training Details}
\noindent\textbf{Datasets.} We performed experiments on two different datasets: AI2D \cite{kembhavi2016diagram} and FOODWEBS \cite{krishnamurthy2016semantic}. AI2D contains approximately 5,000 diagrams representing scientific topics at an elementary school level. 
Overall, the AI2D dataset contains class-annotation for more than 118K constituents and 53K relationships among them, including segmentation mask for each of the elements. AI2D also contains more than 15,000 multiple choice questions about diagrams. 
The polygons for segmentation provided with the AI2D dataset were reshaped into rectangles for simplicity and efficiency. 
FOODWEBS consists of 490 food web diagrams and 5,208 questions encountered on eighth grade science exams. FOODWEBS focuses on question answering using questions about environmental problems. Unlike AI2D, the diagrams in FOODWEBS do not have annotations for relations among objects, and we used this dataset only as a benchmark of question answering.
\\

\noindent\textbf{Baseline.} we used the following ablation models to compare with our method:

\begin{itemize}
\item Fully connected layer - only incorporating the %first
object detection branch in our model and replacing the %second 
graph generation branch with fully connected layers.  
\item Vanilla GRU - similar to the previous baseline but using a vanilla GRU instead of the graph generation branch.
\item DGGN w/o global feature - exploiting the same structure as our model but excluding the global feature from inputs in the second branch. 
\item DGGN w/o weighted mean pool - averaging hidden vectors of adjacent edges without multiplying weights which represent the strength of each adjacency.
\item DGGN w/ ROI-pooled feature - concatenating a $2 \times 2$ ROI-pooled feature in the local feature $f$, expanding it into a 34 dimensional vector.
\end{itemize}

\noindent\textbf{Metrics.}
%nspired by evaluation metrics in object detection, and previous works, 
We propose to measure mean Average Precision (AP) for edge evaluation and IoU for graph completion. First, AP can measure both the recall and precision of a model in predicting the existence of edges. Since our relation candidate should have two boxes, we use average IoUs of those boxes with ground truth boxes as IoU for a relation. 
%A predicted label of a candidate was considered to be true positive if and only if: 1) IoU is higher than threshold, 2) IoU is the highest among 1).
We used IoU thresholds $\tau \in \{0.3, 0.4, 0.5, 0.6, 0.7\}$ for experiments and report the mean AP by averaging the results of all the thresholds.

Additionally, we adopt an IoU metric to measure completion of entire graph. For both of nodes and edges, we define IoU of node and edge as the number of the intersection divided by the number of the union. Note that we only use the number of overlapped nodes or edges instead of using overlapped area in the original IoU metric.
\\

\begin{table}
\begin{center}
\caption{Comparison results of AP on the AI2D test set.}
%Top: results of models \js{using alternative} methods in \js{the} second branch. Middle: results of ablation models of DGGN.
\label{table:tab1}
\resizebox{0.99\linewidth}{!}{
\begin{tabular}{l|cccc}
\hline
Method & $mAP$ & $AP_{30}$ & $AP_{50}$ & $AP_{70}$  \\
\hline\hline
Fully connected layer & 8.87 & 9.22 & 8.92 & 8.24 \\
Vanilla GRU & 39.28 & 39.89 & 43.11 & 31.54 \\
\hline
DGGN \\
\hspace{2pt} w/o global feature & 39.34 & 40.51 & 43.03 & 31.11 \\
\hspace{2pt} w/o weighted mean pool& 42.15 & 44.22 & 44.99 & 34.37 \\
\hspace{2pt} w/ ROI-pooled feature & 39.73 & 43.09 & 42.19 & 31.38 \\
\hline\hline
\textbf{DGGN} & \textbf{44.08} & \textbf{44.23} & \textbf{47.13} & \textbf{38.97} \\
\hline
\end{tabular}}
\end{center}
\vspace{-3mm}
\end{table}

\noindent\textbf{Implementation Details.} %We jointly optimized overall loss with ADAM optimizer with default parameters ($\beta_2 = 0.999, \epsilon = 10^{-9}$). For three losses in overall loss(5), we set weights as $\alpha = 0.2$, $\beta = 0.1$ and $\gamma = 1.0$. The initial learning rate is set to $1\times \epsilon^{-4}$ and is multiplied by 0.09 in everay 1000 iterations. The batch size is set to 32 and we evaluate our model after 15000 iteration ($\approx $150 epochs). 
%Our implementation of the first branch is based on SSD V2 modified from the original SSD. 
We implemented the first branch based on SSDv2 modified from the original SSD.
%In advance, we have fine-tuned the first branch with object detection boxes roughly obtained from segmentation annotations in AI2d dataset. 
%In the second branch, we use 1 layer GRU with 128 hidden states. During training, we sample 160 positive and negative relationship candidates at a ratio of 1 to 7. 
For the second branch, we use 1 layer GRU with 128 hidden states. During training, we sample 160 positive and negative relationship candidates at a ratio of 1 to 7.
The training and testing codes are built on Pytorch. Additional experiments about QA on diagrams utilized the implementation\footnote{https://github.com/allenai/dqa-net} under the same conditions of previous work \cite{kembhavi2016diagram}.

\begin{table}
\begin{center}
\caption{Comparison results of IoU on the AI2D test set.}
\label{table:tab2}
\begin{tabular}{l|cc}
\hline
Method & $IoU_{node}$ & $IoU_{edge}$ \\
\hline\hline
Vanilla GRU & 70.06 & 15.58 \\
\hline
DGGN \\
\hspace{2pt} w/o global feature  & \textbf{70.95} & 14.44 \\
\hspace{2pt} w/o weighted mean pool& 69.48 & 24.84 \\
\hspace{2pt} w/ ROI-pooled feature & 69.24 & 23.00 \\
\hline\hline
\textbf{DGGN} & 69.77 & \textbf{25.86} \\
\hline
\end{tabular}
\end{center}
\vspace{-2mm}
\end{table}

\begin{table}
\begin{center}
\caption{Accuracy of Question Answering on AI2D and FOODWEBS. The results of VQA and DQA-Net(Dsdp) on AI2D and FOODWEBS are refer to \cite{kembhavi2016diagram} and \cite{krishnamurthy2016semantic}, respectively.}
\label{table:tab3}
\begin{tabular}{lcc}
\hline
Method & AI2D \cite{kembhavi2016diagram}  & FOODWEBS \cite{krishnamurthy2016semantic}\\
\hline\hline
Dqa-Net(GT) & 41.55 & - \\
\hline
VQA & 32.90 & 56.50 \\
Dqa-Net(Dsdp) & 38.47 & \textbf{59.30} \\
\textbf{Dqa-Net(Ours)} & \textbf{39.73} & 58.22 \\
\hline
\end{tabular}
\end{center}
\vspace{-2mm}
\end{table}

\subsection{Quantitative Results}

Table \ref{table:tab1} shows comparisons DGGN with baselines on the AI2D dataset. Our results demonstrate that the \textit{DGGN} outperforms baselines. In the second row of the table \ref{table:tab1}, the \textit{Fully connected layer} model shows 8.87 mAP, which is extremely low.
This is because the relational information among the nodes (elements) is not reflected to fully connected layer.
The \textit{vanilla GRU} shows 39.28 mAP, which is lower than those of any variants of \textit{DGGN}. 
This implies that the vanllia GRU model was not successful for embedding the relational information among the nodes, because the GRU model can only learn the sequential order of the input.
In this problem, however, the shuffled order of the relation candidates does not have meaningful sequential knowledge of the relationship.
%, as confirmed by other APs.

Next, we performed ablation studies with variants of \textit{DGGN} as presented in the middle of Table \ref{table:tab1}. 
In the table, we can see that \textit{DGGN w/o global feature} achieved the largest margin to the best model, and this indicates that the global feature can significantly enhance the performance. On the other hand, the result of \textit{DGGN w/o weighted mean pool} is slightly lower than the best model which shows that weights might not be meaningful to the performance.
Interestingly, \textit{DGGN w/ ROI-pooled feature} scored a lower mAP in spite of the additional information. One possible reason is that ROI-pooled feature can cause overfit without a sufficient amount of training data, since objects in diagrams are hard to be generalized.

%In the bottom of Table \ref{table:tab1}, we compare two weight selection methods in retrieve step of DGGN as soft and hard selection. Soft selection method corresponds to our proposed model. To implement hard selection, we generate a mask which draws binary random numbers from a bernoulli distribution. With a mask, weak tie of graph can be removed. As a result, soft selection model outperform hard selection model. Thus, messages from weak tie can contribute to construct a graph.  

Table \ref{table:tab2} shows comparisons of the modified $IoU$ metric for measuring completion of a graph. In the case of the edge inference, we set 0.5 as the threshold of mean IoU of each predicted box intersecting with a ground truth box and set 0.01 as the threshold of confidence for the adjacency of edges.
Since all models use the same SSD model at the object detection branch, results of the $IoU_{node}$ are similar to each other. They have slight different performance because of the end-to-end fine-tuning process.
For $IoU_{edge}$, the \textit{DGGN} shows a better performance than other baselines. Like the results of mAP in Table \ref{table:tab1}, the usage of global feature has a significant impact on the performance.

%As shown in Table 3, we conducted further question answering experiments on AI2D and FOODWEBS. 
Table~\ref{table:tab3} shows the results of the question answering experiments conducted on AI2D and FOODWEBS.
For AI2D, we first evaluated Dqa-net with ground truth annotations of diagrams as our upper bound. 
Our model shows an accuracy of 39.73\% which outperforms previous work and approaches upper bound by 2 \% margin. 
On FOODWEBS, we only deploy on trained model with AI2D and extract diagram graphs from entire data. 
The results show our model demonstrates comparative results. 
Overall, our model performs better when compared with the VQA method, which estimates the answer directly from a diagram image. 
These question answering tests reveal a potential for expansion to the linguistic field.
Also, this result is meaningful in that our model is not directly designed to solve the QA problem.

\subsection{Qualitative Results}

In this section, we analyze qualitative results as shown in Figure \ref{fig:fig7}.
Three diagrams which have different layouts and topics are presented to compare qualitatively in Figure \ref{fig:fig7}(a). For example, diagrams for the same topic ``life cycle of a ladybug" in the second and third row have different layouts. Nevertheless, our model can understand different layouts and generate graphs according to the intentions of the diagrams.
In the second column, the detection results of the object detection branch, finding four kinds of objects (blob, text, arrow and arrow head) in the diagram, are presented.
In the third column, we present the results of graph generation on various diagrams.
Then we compare our results to those of the baseline (vanilla GRU) as shown in the last column.
As seen in the results, we confirmed that our model correctly connected the links between the objects according to their intended relation, in most case.

Figure \ref{fig:fig7}(b) shows a sample describing a pipeline of solving question-answering from a diagram graph. 
After the diagram of ``life cycle of a frog" is converted to a relation graph, we can generate knowledge sentences such as ``Adult Frog links to Young Frog" with three categories : ``relation", ``entity" and ``count".
Using those sentences, we solved the multi-choice QA problems. 
For instance, 
%the first question ``The diagram depicts the life cycle of" asks for a word to complete the question sentence properly. 
%QA model can choose ``a) frog" with a confidence of 0.69 from among the answer options, since we extract the word ``frog" from the diagram. 
the second question asks for the relationship among the objects in the diagram. 
We have already generated a knowledge sentence ``Egg Mass links to Tadpole", so the QA model can easily respond ``b) tadpole" with a confidence of 0.82. 
This process can contribute to the solution of various problems related to knowledge of relationships.

\begin{figure}[t]
\begin{center}
   \includegraphics[width=\linewidth]{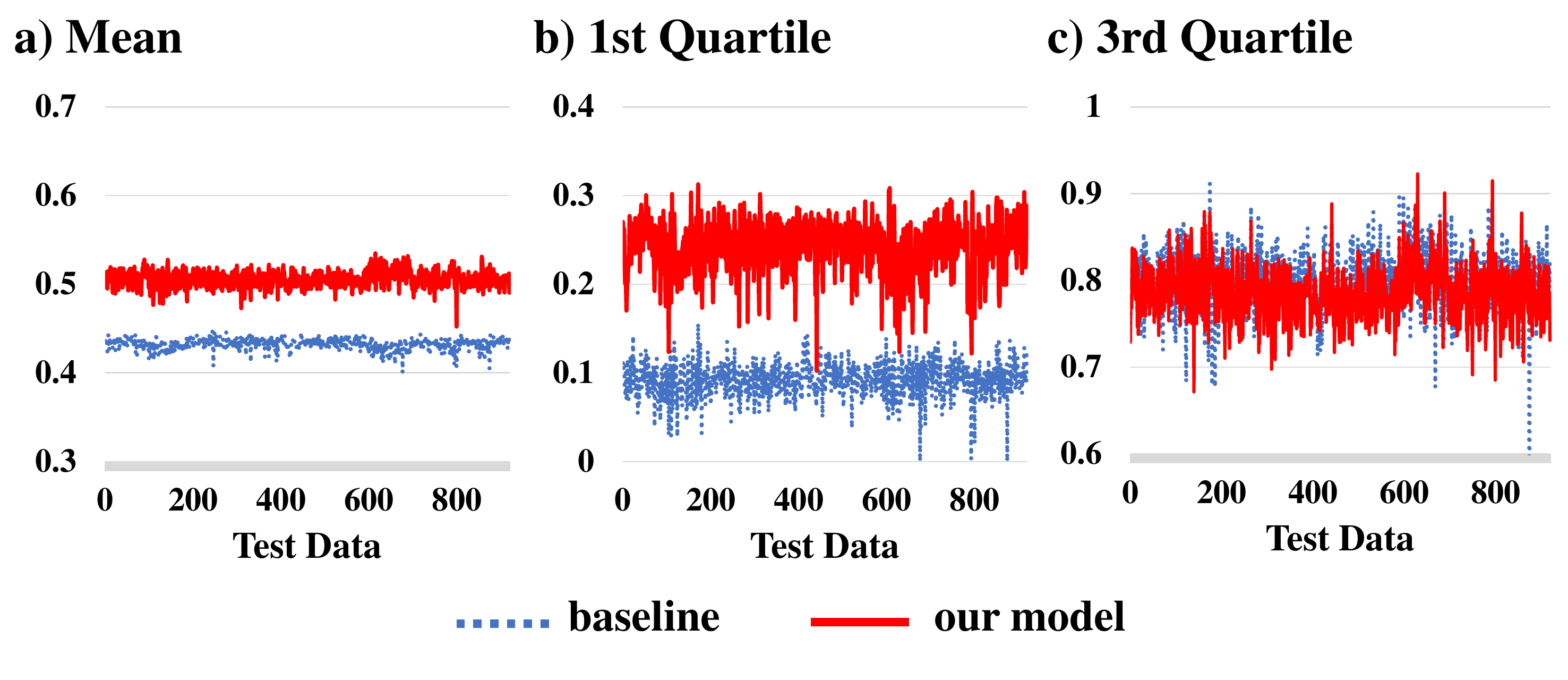}
\end{center}
\vspace{-2mm}
   \caption{Three statistics of activation value of update gate on AI2D test sets. (a) Mean of activation values. (b) The first quartile statistics of activation values. (c) The third quartile statistics of activation values.
   }
\label{fig:fig5}

\end{figure}

\section{Discussion}
\label{sec:discussion}

%\subsection{Analysis of DGGN}

In this section, we discuss the effectiveness of DGGN by investigating the GRU cells, and we analyzed the 
dependency of candidate order of DGGN to compare the 
results between our model and baseline (vanilla GRU).
\\

\noindent\textbf{Activation of gates.} To understand the DGGN better, we analyze information dynamics in DGGN. 
For this, we extracted the activation values of the update gate.
In equation (\ref{eq:gru4}), update gate $z_t$ obtained from equation (\ref{eq:gru2}) %can 
determines the amount of the received information of the cell from the previous $\hat{h}_{t-1}$. 
By investigating the graph of the update gate's activation, we can observe that this model meaningfully exploits messages from the past. 
Obviously, the more update gates activate, the richer the transmitted information becomes. 

We plot three statistics of activation values of update gates using 920 test samples. 
In Figure \ref{fig:fig5}(a), we presented the mean of activation values which shows the significant margin between our model (red solid line) and the baseline (blue dotted line), and this shows that our model can generally activate update gates more effectively than the baseline does.
While the first quartile statistics in Figure \ref{fig:fig5}(b) show a larger margin than the an aforementioned result, the third quartile statistics  
do not show meaningful differences between our model and the baseline in Figure \ref{fig:fig5}(c).
Those two results show that our model encouraged activation in relatively inactive update gates.
Overall, we can conclude that DGGN delivers more informative messages based on the graphical structure to GRU cells and induces more influx of information from the past which can lead to better results.

\begin{figure}[t]
\begin{center}
   \includegraphics[width=\linewidth]{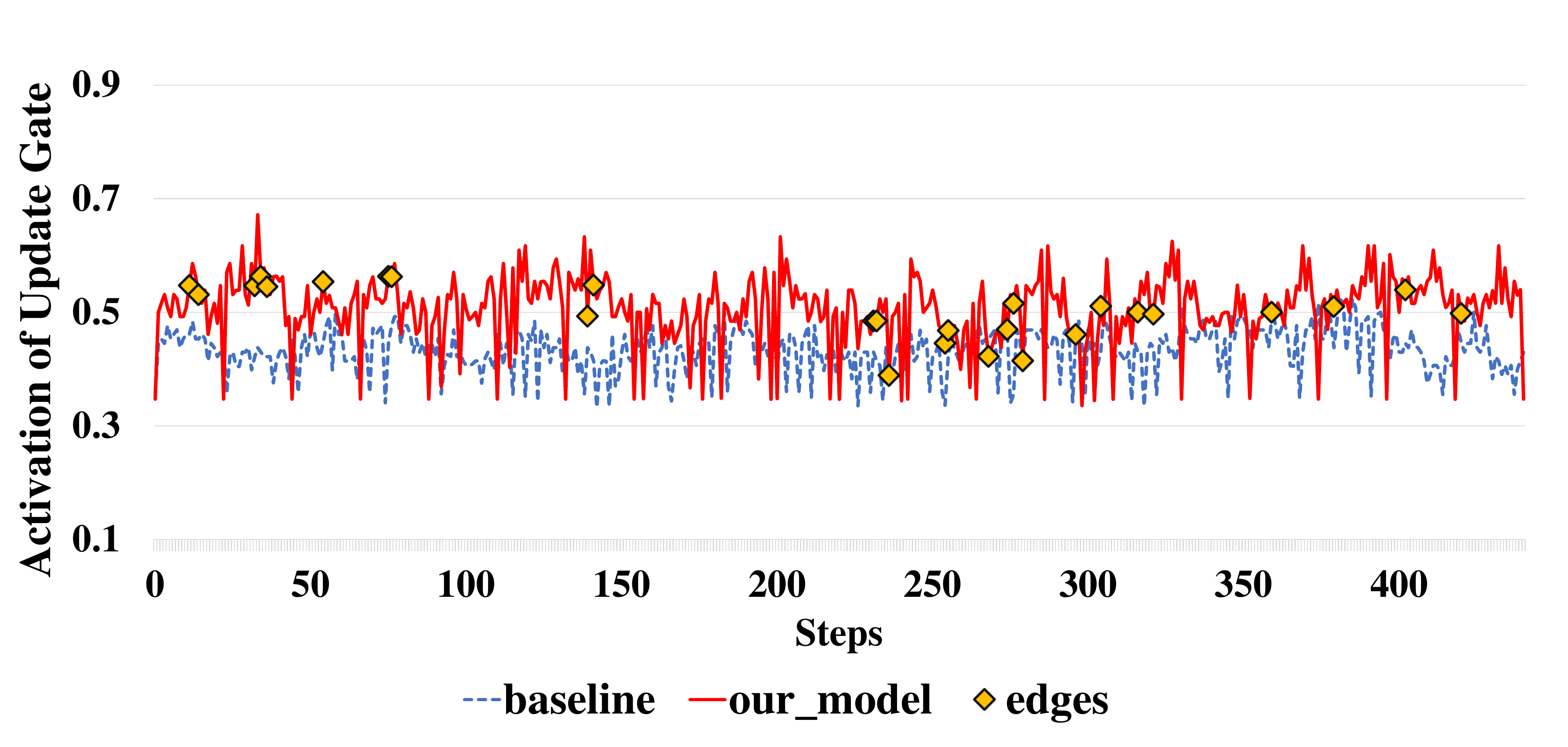}
\end{center}
\vspace{-2mm}
   \caption{Mean of activation values of update gate on second diagram of Figure \ref{fig:fig7}. }
\label{fig:fig6}
\end{figure}

For a study in terms of time steps, we extract activation values of update gates in GRU cells from the second diagram in Figure \ref{fig:fig7}(a). 
Then we average this quantity over hidden cells with respect to time steps. 
As shown in Figure \ref{fig:fig6}, our model performs higher than the baseline over almost all the steps. 
Specifically, almost all the yellow dots in the graph, depicting the candidates of the connected edge, show that the activation values of our model are higher than baseline.
Therefore, as we discussed in the previous chapter, the cells of our model successfully infer the relationships by accepting more adjacent information with respect to time steps. 
\\

\noindent\textbf{Order of relation candidates. }   To explore a mechanism of aggregating messages in DGGN, we verified the effect of the order of relation candidates. 
We evaluated 50 results ($AP_{50}$) repeatedly with randomly ordered candidates for the baseline and our model on the AI2D. Then we extracted variation statistics from the results. 
For the baseline, variance and standard deviation of results are $2.27e^{-5}$ and $4.76e^{-3}$, respectively.
Our model shows a variance and a standard deviation of results of $1.03e^{-7}$ and $3.22e^{-4}$, respectively. 
The result shows that the variance and the standard deviation of our model are much lower (around 13 times smaller standard deviation) than those of the baseline.

During the training process, we shuffled the order of candidates before transmitted into GRU cells for both models, to avoid order dependency. 
However, the statistics show that our model is more robust against the order of relation candidates compared to the baseline. 
We can confirm that the proposed model successfully extracts the graph structure regardless of the order of the input sequence, due to the proposed method's ability to aggregate messages from the past. 

\section{Conclusion}
\label{sec:conclusion}

In this work, we proposed \textit{UDPnet} and \textit{DGGN} to tackle the problem of understanding a diagram and generating a graph by the neural network. For diagram understanding, we combine an object detector and a network that generates relations among detected objects.
A multi-task learning scheme is used to train the UDPnet in an end-to-end manner. Moreover, we propose a novel RNN module to propagate message based on graph structure and generate a graph simultaneously. Then, we demonstrated that the proposed UDPnet provides state-of-the-art quantitative and qualitative results on problems of generating relation for a given diagram.
We also analyzed how our model works better than strong baselines. Our work can be a meaningful step in diagram understanding and reasoning problem beyond natural image understanding. Moreover, we believe that the proposed DGGN could benefit other tasks related to graph structure.

{\small
\bibliographystyle{ieee}
\bibliography{egbib}
}

\clearpage
\onecolumn

%\begin{center}
%\textbf{\large Appendix}
%\end{center}

%\setcounter{equation}{0}
%\setcounter{figure}{0}
%\setcounter{table}{0}
%\setcounter{page}{1}
%\setcounter{section}{0}

\section{Appendix}

\subsection{Results of Recall@K}

While we measured AP for evaluation in this paper, we additionally utilized recall metric for measuring retrieval power of relationships due to the sparsity of the relationship. Table \ref{table:sup_tab1} shows results of Recall@k metric (R@k) on AI2D test dataset. The R@k measures the fraction of ground-truth relationship that appears among the top-k most confident predictions. 
The results of R@5, R@10 and R@20 demonstrate similar trend of results compared to those of mAPs. 

\begin{table}[h]
\begin{center}
\caption{Comparison results of Recall@K on the AI2D test set.}
\label{table:sup_tab1}
\resizebox{0.5\linewidth}{!}{
\begin{tabular}{l|ccc}
\hline
Method & $R@5$ & $R@10$ & $R@20$  \\ %& $R@50$
\hline\hline
Vanilla GRU & 21.79 & 33.97 & 48.87  \\ % & 66.16
\hline
DGGN \\
\hspace{2pt} w/o global feature & 21.44 & 33.63 & 49.18  \\ % & \textbf{68.34}
\hspace{2pt} w/o weighted mean pool& 22.62 & 35.75 & 51.60  \\ % & 68.10
\hspace{2pt} w/ ROI-pooled feature & 21.45 & 34.21 & 49.87   \\ % & 67.15
\hline\hline
\textbf{DGGN} & \textbf{22.66} & \textbf{35.93} & \textbf{51.73}   \\ % & 68.16
\hline
\end{tabular}}
\end{center}
\end{table}

\subsection{Training Details}

For training, We jointly optimized the overall loss of the proposed algorithm with ADAM optimizer with default parameters ($\beta_2 = 0.999, \epsilon = 10^{-9}$). 
For the three losses in overall loss (\ref{eq:loss_sup}), we set %weights as 
$\alpha = 0.2$, $\beta = 0.1$ and $\gamma = 1.0$. 
The initial learning rate is set to $1\times \epsilon^{-4}$ and is multiplied by 0.09 in every 1000 iterations. 
The batch size is set to 32 and we evaluated our model after 15000 iteration ($\approx $150 epochs). 

\begin{equation} 
L = \alpha L_c + \beta L_l + \gamma L_r.
\label{eq:loss_sup}
\end{equation}

\subsection{Details of Post-processing}

\begin{algorithm}
\caption{Post processing algorithm}\label{postpro}
\begin{algorithmic}[1]
\Require
Relation set $R$ generated by the proposed DGGN
\Ensure Generated sentences set $S$
\State $S \gets \emptyset$ 
\Repeat
\State $R_a \gets \{o_{a1}, o_{a2}\}\in R$
\State $R_b \gets \{o_{b1}, o_{b2}\}\in R$
\If{ $R_a \cap R_b$ $\in$ `$text$' }
    \State Continue
\ElsIf{$R_a \cap R_b$ $\in$ `$blob$'}
    \If{$R_a - R_b $ $\in$ `$text$' \& $R_b - R_a $ $\in$ `$text$'}
        \State{Generate sentence $S_{ab}$}
        \State $S \gets S \cup S_{ab}$
    \ElsIf{$R_a - R_b $ $\in$ `$text$' \& $R_b - R_a $ $\in$ `$blob$'}
        \State Find $R_c$ satisfying \{$R_c\cap R_b \in `blob$' \& $R_c - R_b \in `text$' \}
        \State{Generate sentence $S_{ac}$}
        \State $S \gets S \cup S_{ac}$

    \EndIf

\EndIf
\Until{all elements in $R$ are visited}
\end{algorithmic}
\end{algorithm}

In this section, we explain a detailed post-processing procedure of the proposed method.
Once relationships are determined among objects, we can additionally make %further 
new relationship between objects sharing the same intermediate node. 
For example, given two text objects sharing the same blob object, we can say that one text is linked to another one. 
Also, given two text objects connected by two intermediate blob objects, we can say %the same as
equivalently to the previous case. 
In most case, %Usually  
the text object represents the name or explanation of the connected blob object. 
Consequently, making further connections by this rule-based algorithm, we can generate sentences using given texts. Using extensively connected texts, we just put an additional phrase of ``links to" such as ``Lavar links to Fly".
Note that localized text boxes are recognized using Tesseract\footnote{https://github.com/tesseract-ocr/tesseract}.
Algorithm 1 shows details of post processing.

After generating sentences about relationship, we added sentences about facts of detected elements in a diagram. As shown in Figure \ref{fig:sup4}, counts of objects and stages, and names of elements are added to help richer descriptions about a diagram.

\subsection{Additional Qualitative Results}

\begin{figure*}[t]
\begin{center}
\includegraphics[width=\textwidth]{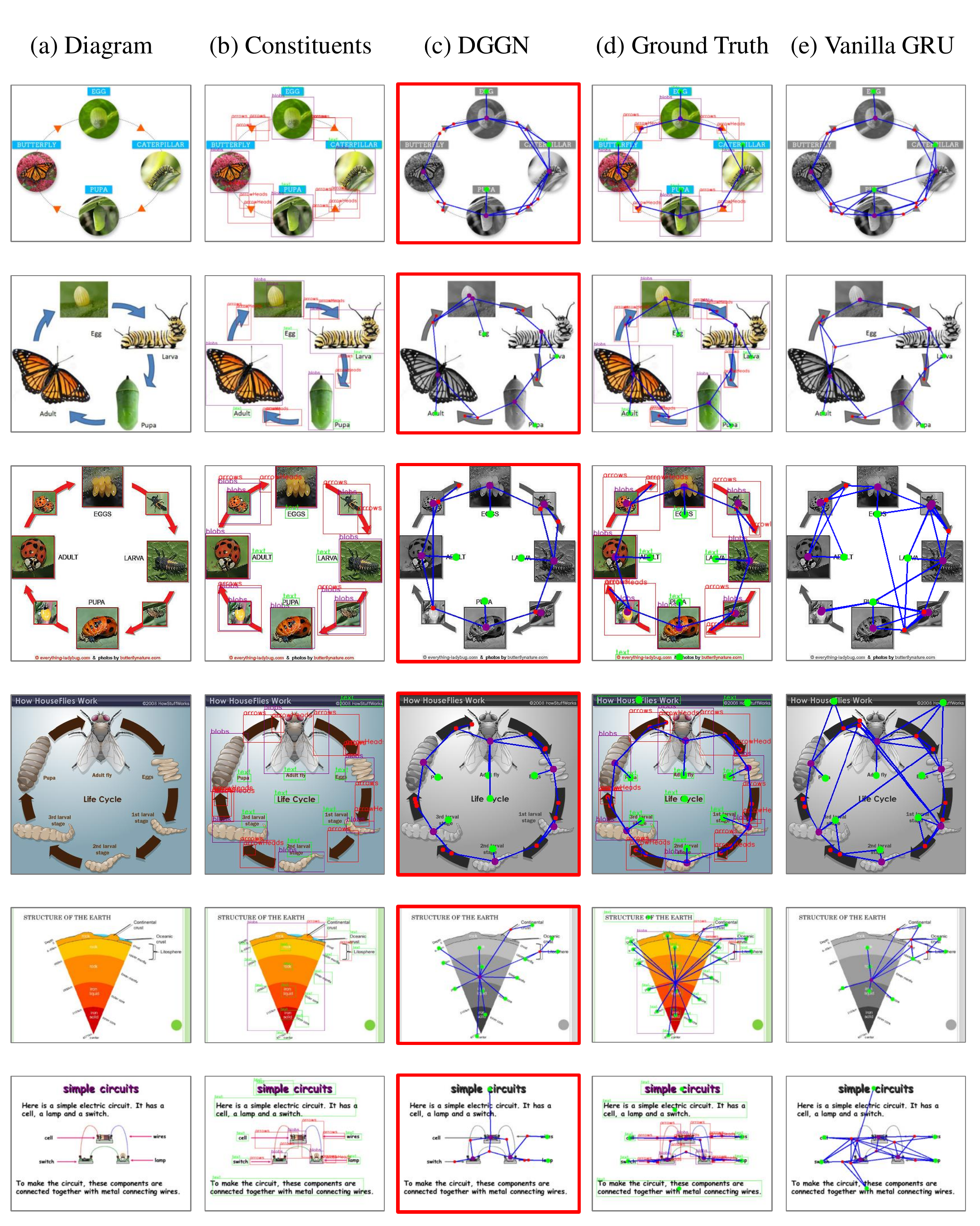}
\end{center}
   \caption{Additional qualitative results on diagram graph generation: (a) original diagram (b) diagram with detected constituents (c) generated graph results of DGGN (d) ground truth (e) results of baseline (vanilla GRU)}
\label{fig:sup1}
\end{figure*}

\begin{figure*}[t]
\begin{center}
\includegraphics[width=\textwidth]{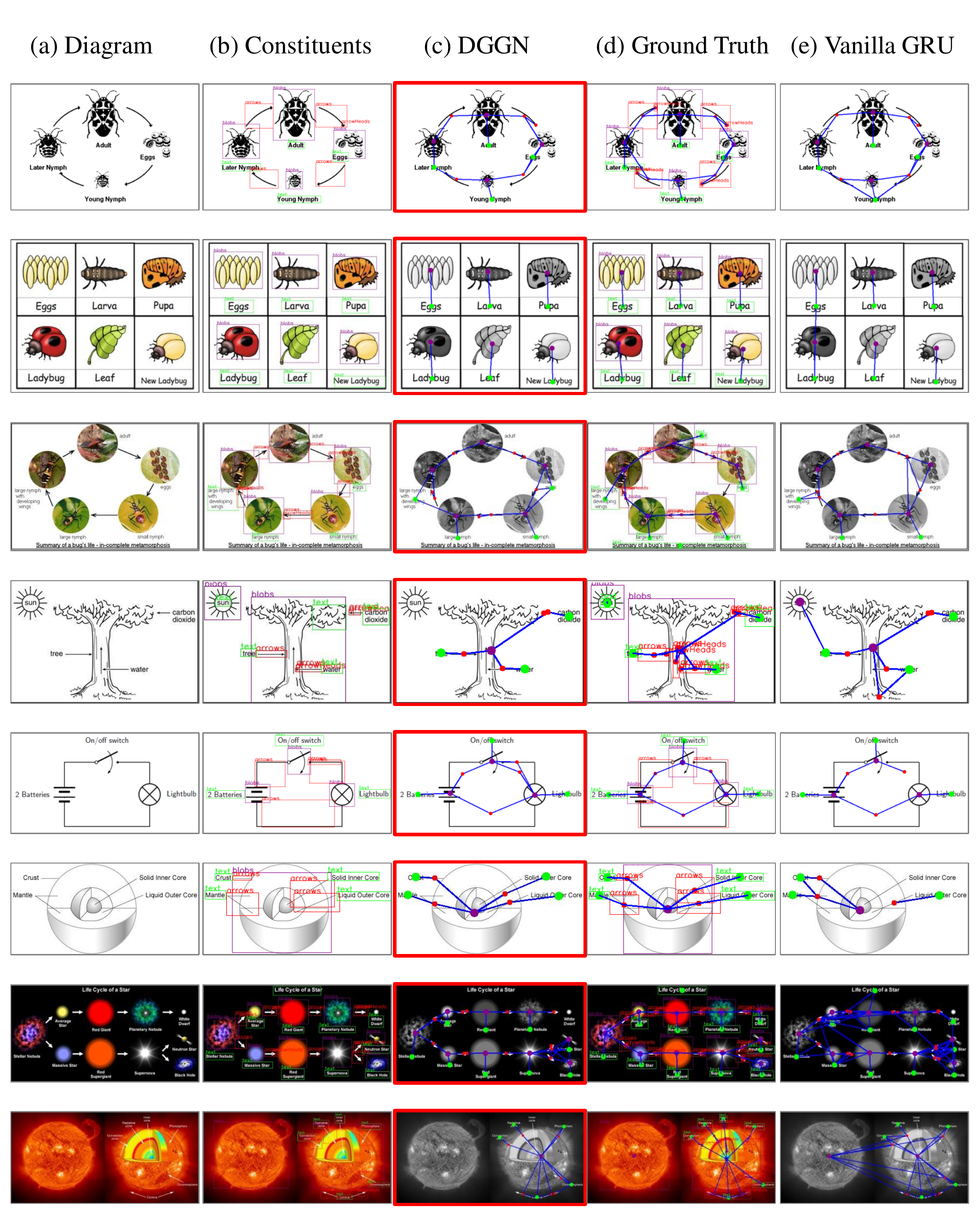}
\end{center}
   \caption{Additional qualitative results on diagram graph generation: (a) original diagram (b) diagram with detected constituents (c) generated graph results of DGGN (d) ground truth (e) results of baseline (vanilla GRU)}
\label{fig:sup2}
\end{figure*}

\begin{figure*}[t]
\begin{center}
\includegraphics[width=\textwidth]{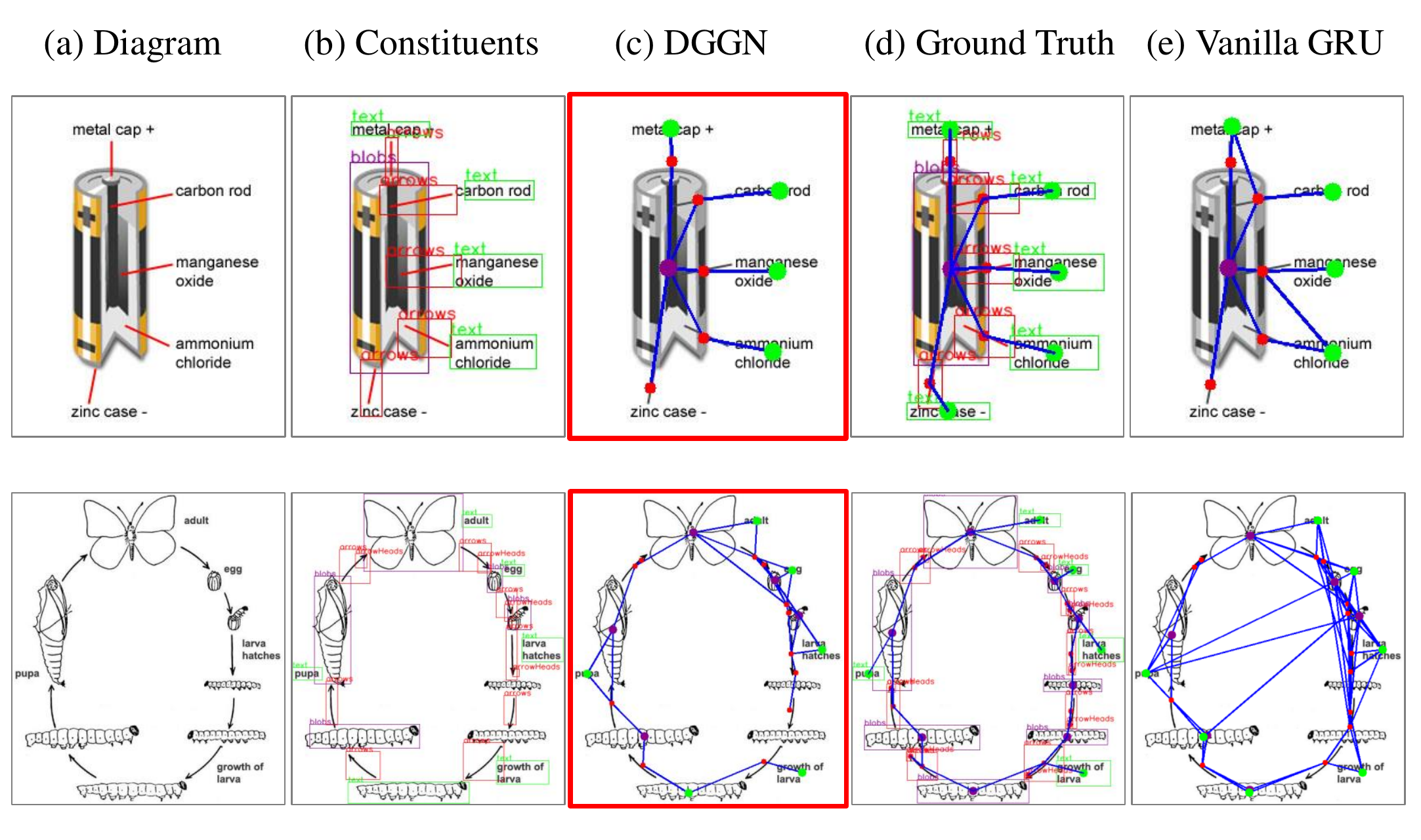}
\end{center}
   \caption{Additional qualitative results on diagram graph generation: (a) original diagram (b) diagram with detected constituents (c) generated graph results of DGGN (d) ground truth (e) results of baseline (vanilla GRU)}
\label{fig:sup3}
\end{figure*}

\begin{figure*}[t]
\begin{center}
\includegraphics[width=\textwidth]{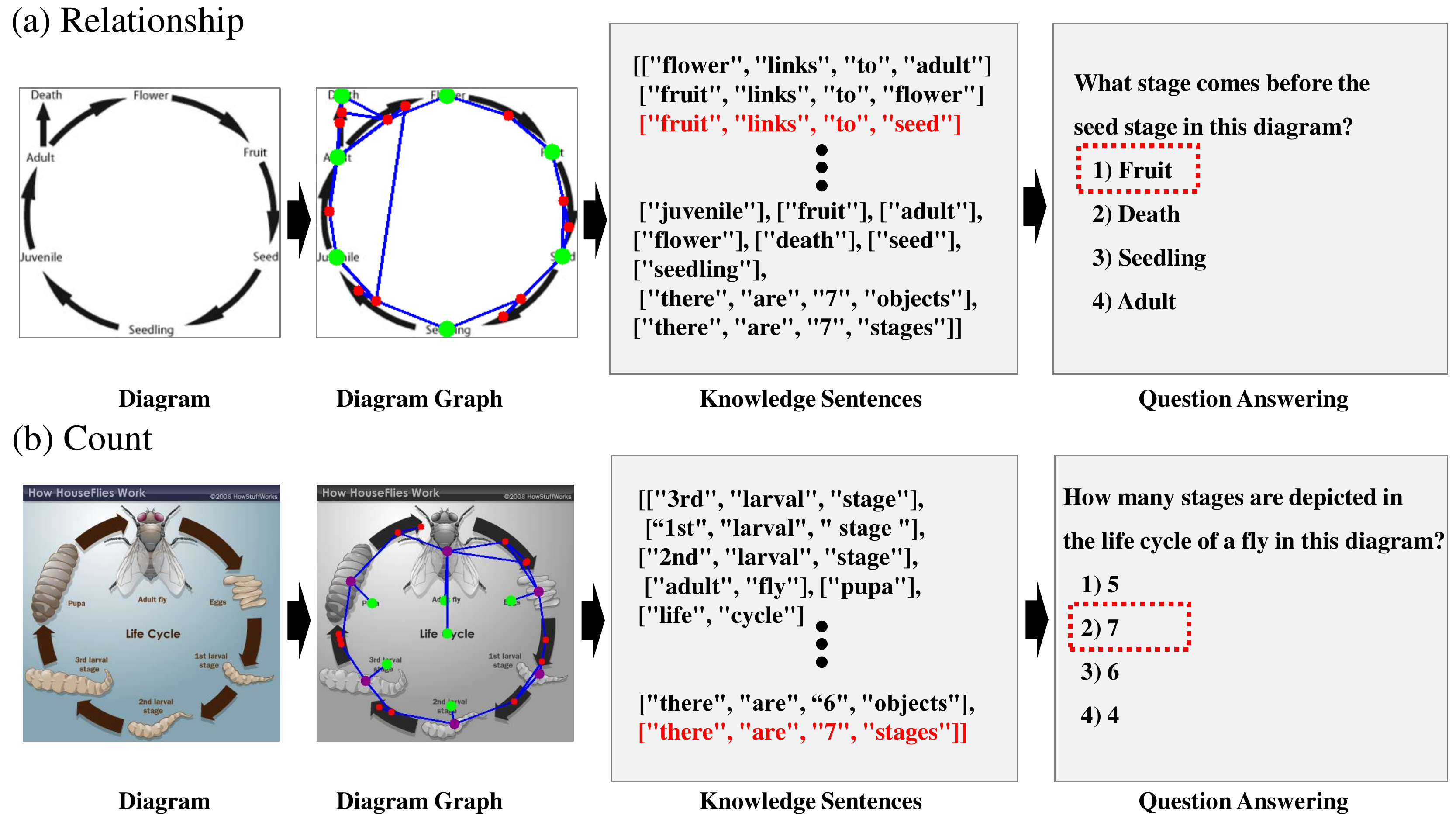}
\end{center}
   \caption{Additional qualitative results on question answering: (a) a question about relationship. (b) a question about the count of stages}
\label{fig:sup4}
\end{figure*}

In next pages, we present additional qualitative results on diagram graph generation and question answering. We provide results of diagram graph generation of various layouts and topics as depicted from Figure \ref{fig:sup1} to Figure \ref{fig:sup3}. The results of DGGN are compared with those of those of vanilla GRU. Ground truths are also shown.
In Figure \ref{fig:sup4}, we also show pipelines from diagram graph to question answering with post-processing described in the previous section. For two different types of questions, \textit{relationship} and \textit{count}, related sentences are highlighted.

%{\small
%\bibliographystyle{ieee}
%\bibliography{egbib_sup}
%}

\end{document}